%% file: main.tex
\documentclass[sigconf]{acmart}

\AtBeginDocument{%
  \providecommand\BibTeX{{%
    \normalfont B\kern-0.5em{\scshape i\kern-0.25em b}\kern-0.8em\TeX}}}

\copyrightyear{2020} 
\acmYear{2020} 
\setcopyright{acmcopyright}\acmConference[KDD '20]{Proceedings of the 26th ACM SIGKDD Conference on Knowledge Discovery and Data Mining}{August 23--27, 2020}{Virtual Event, CA, USA}
\acmBooktitle{Proceedings of the 26th ACM SIGKDD Conference on Knowledge Discovery and Data Mining (KDD '20), August 23--27, 2020, Virtual Event, CA, USA}
\acmPrice{15.00}
\acmDOI{10.1145/3394486.3403308}
\acmISBN{978-1-4503-7998-4/20/08}

\usepackage{algorithm,algpseudocode,color,graphicx,multirow,paralist,pifont,subfigure,url,xspace,multirow}
\usepackage{balance}
\usepackage[rightcaption,wide]{sidecap}

\newcommand*{\Scale}[2][4]{\scalebox{#1}{$#2$}}

\begin{document}

\title{Calendar Graph Neural Networks for Modeling Time Structures in Spatiotemporal User Behaviors}
\fancyhead{}

\author{Daheng Wang$^1$, Meng Jiang$^1$, Munira Syed$^1$, Oliver Conway$^2$, Vishal Juneja$^2$}
\author{Sriram Subramanian$^2$, Nitesh V. Chawla$^{1,3}$}
\affiliation{
	\institution{$^1$University of Notre Dame, Notre Dame, IN 46556, USA}
	\institution{$^2$Cond\'{e} Nast, New York, NY 10007, USA}
	\institution{$^3$Department of Computational Intelligence, Wroc\l{}aw University of Science and Technology, Wroc\l{}aw, Poland}
}
\email{{dwang8, mjiang2, msyed2, nchawla}@nd.edu}
\email{{oliver_conway, vishal_juneja, sriram_subramanian}@condenast.com}


\begin{abstract}
\input{0abstract}
\end{abstract}

%


\settopmatter{printacmref=false, printfolios=false}

\maketitle

{\fontsize{8pt}{8pt} \selectfont \textbf{ACM Reference Format:}\\
Daheng Wang, Meng Jiang, Munira Syed, Oliver Conway, Vishal Juneja, Sriram Subramanian, Nitesh V. Chawla. 2020. Calendar Graph Neural Networks for Modeling Time Structures in Spatiotemporal User Behaviors. \textit{In The 26th ACM SIGKDD Conference on Knowledge Discovery \& Data Mining (KDD '20), August 23--27, 2020, Virtual Event, CA, USA.} ACM, NY, NY, USA, 9 pages. https://doi.org/10.1145/3394486.3403308}

\section{Introduction}
\label{sec:introduction}
\input{1introduction}

\vspace{-0.1in}
\section{Related Work}
\label{sec:related}
\input{2related}

\section{Problem Definition}
\label{sec:problem}
\input{3problem}

\section{The CalendarGNN Framework}
\label{sec:approach}
\input{4approach}

\section{Experiments}
\label{sec:experiments}
\input{5experiments}

\section{Conclusions}
\label{sec:conclusions}
\input{6conclusions}

\begin{acks}
This research was supported in part by Cond\'{e} Nast, and by NSF Grants IIS-1849816 and IIS-1447795.
This research was also supported in part by the National Science Centre, Poland research project no.2016/23/B/ST6/01735.
\end{acks}

\bibliographystyle{ACM-Reference-Format}
\bibliography{main}


\end{document}

%% file: 0abstract.tex
User behavior modeling is important for industrial applications such as demographic attribute prediction, content recommendation, and target advertising.
Existing methods represent behavior log as a sequence of adopted items and find sequential patterns; however, concrete location and time information in the behavior log, reflecting dynamic and periodic patterns, joint with the spatial dimension, can be useful for modeling users and predicting their characteristics.
In this work, we propose a novel model based on graph neural networks for learning user representations from spatiotemporal behavior data.
Our model's architecture incorporates two networked structures. One is a tripartite network of items, sessions, and locations. The other is a hierarchical calendar network of hour, week, and weekday nodes.
It first aggregates embeddings of location and items into session embeddings via the tripartite network, and then generates user embeddings from the session embeddings via the calendar structure. The user embeddings preserve spatial patterns and temporal patterns of a variety of periodicity (e.g., hourly, weekly, and weekday patterns).
It adopts the attention mechanism to model complex interactions among the multiple patterns in user behaviors.
Experiments on real datasets (i.e., clicks on news articles in a mobile app) show our approach outperforms strong baselines for predicting missing demographic attributes.


%% file: 1introduction.tex
\begin{figure}[t]
	\centering
	\vspace{0.2in}
	{\includegraphics[width=1\columnwidth]{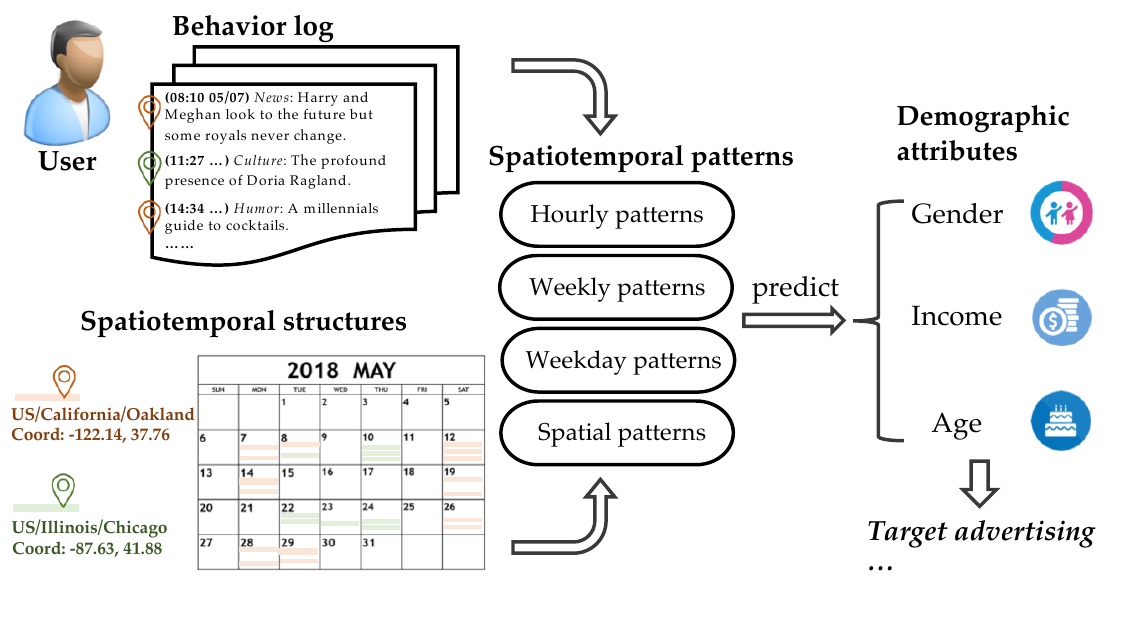}}
	\vspace{-0.2in}
	\caption{Our framework incorporates calendar structure to model spatiotemporal patterns (including multi-level periodicity) for predicting missing demographic attributes.}
	\vspace{-0.15in}
	\label{fig:intro}
\end{figure}

Online web platforms have large databases to record user behaviors such as reading news articles, posting social media messages, and clicking ads. Behavior modeling is important for a variety of applications such as user categorization \cite{boratto2016using}, content recommendation \cite{wu2017session,li2017neural}, and targeted advertising \cite{aly2012web}. Typical approaches learn users' vector presentations from their behavior log for predicting missing demographic attributes and/or preferred content. 

Spatiotemporal patterns in behavior log are reflecting user characteristics and thus expected to be preserved in the vector representations. Earlier work modeled a user's temporal behaviors as a sequence of his/her adopted items and used recurrent neural networks (RNNs) to learn user embeddings \cite{hidasi2015session}. For example, Hidasi \emph{et al.} proposed parallel RNN models to extract features from (sequential) session structures \cite{hidasi2016parallel}; Tan \emph{et al.} proposed to model temporal shifts in RNNs; and Jannach \emph{et al.} combined RNNs with neighborhood-based methods to capture sequential patterns in user-item co-occurrence \cite{jannach2017recurrent}. Recently, Graph Neural Networks (GNNs) have attracted increasing interests for learning representations from graph structured data \cite{defferrard2016convolutional,kipf2016semi,gilmer2017neural,velivckovic2017graph}. The core idea is to use convolution or aggregation operators to enhance representation learning through the graph structures \cite{bruna2013spectral,hamilton2017inductive,ying2018graph}. For modeling temporal information in network, Manessi \emph{et al.} \cite{manessi2017dynamic} stacked RNN modules \cite{hochreiter1997long} on top of graph convolution networks \cite{kipf2016semi}; Seo \emph{et al.} \cite{seo2018structured} replaced fully connected layers in RNNs with graph convolution \cite{defferrard2016convolutional}. However, existing GNNs can only model sequential patterns or incremental changes in graph series. The spatiotemporal patterns are much more complex in real-world behavior log.

In existing GNN-based user models, the missing yet significant type of patterns is \emph{periodicity} at different levels such as hourly, weekly, and weekday patterns (see Figure \ref{fig:intro}). For example, some users may have the habit of browsing news articles early in the morning during workdays; some may browse news late at midnight right before sleep. To discover these patterns one needs to process concrete time information beyond simple sequential ordering. So the time levels (or say, the hierarchical structure of calendar) must be incorporated into the process of user embedding learning.

User behaviors exhibit temporal patterns across different periodicities on the time dimension. Our idea is to leverage the explicit scheme of the \emph{Calendar} system for modeling the hierarchical time structures of user behaviors. A standard annual calendar system, e.g., the Gregorian calendar, imposes natural temporal units for timekeeping such as day, week, and month. A daily calendar system imposes more refined temporal units such as hour and minute. These temporal units can naturally be applied to frame temporal patterns. Patterns of various periodicity can be complementary with each other when jointly learned to extract user representations.

In this work, we propose a novel GNN-based model, called \textsc{CalendarGNN}, for modeling spatiotemporal patterns in user behaviors by incorporating time structures of the calendar systems as neural network architecture. It has three aspects of  novel designs.

First, a user's behavior log forms a tripartite graph of items, sessions, and locations. In \textsc{CalendarGNN}, session embeddings are aggregated from embeddings of the corresponding items and locations; embeddings of time units (e.g., node ``3PM'', node ``Tuesday'', or node ``the 15th week of 2018'') are aggregated from the session embeddings. The embedding of each time unit captures a certain aspect of the user's temporal patterns. Then the model aggregates these time unit embeddings into temporal patterns of different periodicity such as hourly, weekly, and weekday patterns. The temporal patterns are distilled from all his/her previous sessions happened during the time periods specified by the time unit.

Second, in addition to the temporal dimension, \textsc{CalendarGNN} discovers spatial patterns from spatial signals in user sessions. It aggregates session embeddings into location unit embeddings which can be later aggregated into the user's spatial pattern. The latent user representations are generated by concatenating all temporal patterns and spatial pattern. The user embeddings are used (by classifiers or predictive models) for various downstream tasks.

Third, temporal patterns and spatial patterns should not be separately learned because they interact with each other in user behavior. For example, people may read news at Starbucks in the morning, in restaurants at noon, and at home in the evening; people may prefer different types of topics at different places when they travel to different cities or countries for business. Our model considers the interactions between spatial pattern and the multi-level temporal patterns. We develop a model variant \textsc{CalendarGNN-Attn} that utilizes interactive attentions between location units and different time units for capturing user's complex spatiotemporal patterns.

We conduct experiments on two real-world spatiotemporal behavior datasets (in industry) for predicting user demographic labels (such as gender, age, and income). Results demonstrate the effectiveness of our proposed model compared to existing work.



%% file: 2related.tex
We discuss three lines of research related to our work.

\noindent \textbf{Temporal GNNs.}
The success of \textsc{GNN} on tasks in static setting such as link prediction \cite{ying2018graph, zhang2018link} and node classification \cite{hamilton2017inductive, velivckovic2017graph} motives many work to look at the problem of dynamic graph representation learning.
Some deep graph neural methods explored the idea of combining \textsc{GNN} with recurrent neural network (\textsc{RNN}) for leaning node embeddings in dynamic attributed network \cite{manessi2017dynamic, seo2018structured}.
These methods aim at modeling the structural evolution among a series of graphs and they cannot be directly applied on users' spatiotemporal graphs for generating behavior patterns.
Another set of approaches for spatiotemporal traffic forecasting aim at capturing the evolutionary pattern of node attribute given a fixed graph structure. Li \emph{et al.} \cite{li2017diffusion} modeled the traffic flow as a diffusion process on a directed graph and adopted an encoder-decoder architecture for capturing the temporal attribute dependencies. Yu \emph{et al.} \cite{yu2017spatio} modeled the traffic network as a general graph and employed a fully convolutional structure \cite{defferrard2016convolutional} on time axis. 
These methods assume the graph structure remains static and model the change of node attributes. They are not designed for capturing the complex time structures among a large number of user spatiotemporal graphs.

\noindent \textbf{Graph-level GNNs.}
Different from learning node representations, there are some work focus on the problem of learning graph-level representation leveraging node embeddings.
A basic approach is applying a global sum or average pooling on all extracted node embeddings as the last layer \cite{duvenaud2015convolutional, simonovsky2017dynamic}.
Some methods rely on specifying or learning the order over node embeddings so that CNN-based architectures can be applied \cite{niepert2016learning}.
Zhang \emph{et al.} \cite{zhang2018end} proposed a \textsc{SortPooling} layer to take unordered vertex features as input and outputs sorted graph representation of a fixed size in analogous to sorting continuous \textsc{WL} colors \cite{weisfeiler1968reduction}.
Another way of aggregating node embeddings into graph embedding is learning hierarchical representation through differentiable  pooling \cite{ying2018hierarchical}. 
Simonovsky \emph{et al.} \cite{simonovsky2017dynamic} proposed to perform edge-conditioned convolutions over local graph neighborhoods exploiting edge labels and generate the final graph embedding using a graph coarsening algorithm followed by a global sum pooling layer.
These methods are not designed to model user's spatiotemporal behaviors data and cannot explicitly capture the complex time structures of different periodicity.

\noindent \textbf{Session-based user behavior modeling.}
Hidasi \emph{et al.} \cite{hidasi2016parallel} 
proposed a recurrent neural network based approach for modeling users by employing a ranking loss function for session-based recommendations.
Tan \emph{et al.} \cite{tan2016improved} considered temporal shifts of user behavior \cite{yu2020identifying} and incorporated data augmentation techniques to improve the performance of \textsc{RNN}-based model. Jannach \emph{et al.} \cite{jannach2017recurrent} combined the \textsc{RNN} model with the neighborhood-based method to capture the sequential patterns and co-occurrence signals \cite{jiang2014fema, jiang2016catchtartan}. 
Different from these user behavior modeling methods mostly basing on \textsc{RNN} architectures, our framework models each user's behaviors as a tripartite graph of items, sessions and locations, then learns user latent representations via a calendar neural architecture. One recent work by Wu \emph{et al.} \cite{wu2019session} models user's session of items as graph structure and use \textsc{GNN} to generate node or item embeddings. However, it is not capable of learning user embeddings. Our work aims at learning effective user representations capturing both the spatial pattern and temporal patterns for different predictive tasks.

%% file: 3problem.tex
\begin{table}[t]
	\renewcommand{\arraystretch}{1.1}
	\centering
	\caption{Symbols and their description.}
	\label{tab:notations}
	\vspace{-0.1in}
	\Scale[0.9]{
	\begin{tabular}{|c||l|}
	\hline
		\textbf{Symbol} & \textbf{Description} \\ \hline \hline
		$u$, $s$, $v$, $l$ & a user, a session, an item, and a location \\ \hline
		$\mathcal{U}$, $\mathcal{S}$, $\mathcal{V}$, $\mathcal{L}$ & set of users, sessions, items and locations \\ \hline
		$S$ ($S_u$) & subset of sessions $\mathcal{S}$ of user $u$ \\ \hline
		$V$ ($V_u$) & subset of items $\mathcal{V}$ of user $u$ \\ \hline
		$L$ ($L_u$) & subset of locations $\mathcal{L}$ of user $u$ \\ \hline
		$G_u$ & user $u$'s spatiotemporal behavior graph \\ \hline
		$E$ & edge set of $G_u$ \\ \hline
		$E^{(L)}$ & subset of $E$ containing location-session edges \\ \hline
		$E^{(V)}$ & subset of $E$ containing item-session edges \\ \hline
		$\mathcal{G}$ & set of user spatiotemporal behavior graphs \\ \hline
		$a_u$, $\mathcal{A}$ & user label, and set of user labels \\ \hline
		$\mathcal{B}$ & spatiotemporal behavior graph data \\ \hline
		$\mathbf{u}$, $\mathbf{s}$, $\mathbf{v}$, $\mathbf{l}$ & emb. of user, session, item, and location nodes \\ \hline
		$K_{\mathcal{U}}$, $K_{\mathcal{S}}$, $K_{\mathcal{V}}$, $K_{\mathcal{E}}$ & dimensions of $\mathbf{u}$, $\mathbf{s}$, $\mathbf{v}$, $\mathbf{l}$ vectors \\ \hline
		$h_i$, $w_i$, $y_i$, $l_i$ & hour, week, weekday and location unit of $s_i$ \\ \hline
		$\mathcal{T}_{h}$, $\mathcal{T}_{w}$, $\mathcal{T}_{y}$ & set of temporal units: hour, week, and weekday \\ \hline
		$\mathbf{e}_{h}$, $\mathbf{e}_{w}$, $\mathbf{e}_{y}$, $\mathbf{e}_{l}$ & hour, week, weekday, and location unit emb. \\ \hline
		$\mathbf{p}_{\mathcal{T}_{h}}$, $\mathbf{p}_{\mathcal{T}_{w}}$, $\mathbf{p}_{\mathcal{T}_{y}}$, $\mathbf{p}_{\mathcal{L}}$ & hourly, weekly, weekday, and spatial pattern \\ \hline
		\multirow{2}*{$\mathbf{p}_{\mathcal{T}_{h}}^{\mathcal{L}}$, $\mathbf{p}_{\mathcal{T}_{w}}^{\mathcal{L}}$, $\mathbf{p}_{\mathcal{T}_{y}}^{\mathcal{L}}$} & hourly, weekly, weekday pattern  \\
		 & under impacts from spatial pattern \\ \hline
		 \multirow{2}*{$\mathbf{p}_{\mathcal{L}}^{\mathcal{T}_{h}}$, $\mathbf{p}_{\mathcal{L}}^{\mathcal{T}_{w}}$, $\mathbf{p}_{\mathcal{L}}^{\mathcal{T}_{y}}$} & spatial patterns under impacts  \\
		 & from hourly, weekly, weekday pattern \\ \hline
		 \multirow{2}*{$\mathbf{p}_{\mathcal{L},\mathcal{T}_{h}}$, $\mathbf{p}_{\mathcal{L},\mathcal{T}_{y}}$, $\mathbf{p}_{\mathcal{L},\mathcal{T}_{w}}$} & interactive spatial-hourly,  spatial-weekly \\
		 & and spatial-weekday patterns \\ \hline
	\end{tabular}
	}
	\vspace{-0.15in}
\end{table}

In this section, we first introduce concept of the user spatiotemporal behavior graph then formally define our research problem. The notations used throughout this paper are summarized in Table \ref{tab:notations}.

A traditional online browsing behavior log contains the transaction records between users and the server. Typically, a user can start multiple sessions and each session is associated with one or more items such as news articles or update feeds. For a spatiotemporal behavior log, in addition to the sessions and items information, there are also corresponding spatial information, e.g., the city or the neighborhood, for each session of the user; and, explicit temporal information, e.g., server timestamp, for each item of the session.

\begin{definition}[Spatiotemporal Behavior Log]
\label{def:spatiotemporal_log}
A spatiotemporal behavior log is defined on a set of users $\mathcal{U}$, a set of sessions $\mathcal{S}$, a set of items $\mathcal{V}$, and a set of locations $\mathcal{L}$. For each user $u \in \mathcal{U}$, her behavior log can be represented by a set of session-location tuples $\{(s_{u,1}, l_{u,1}), \dots, (s_{u,m_u}, l_{u,m_u})\}$ where $m_u$ denotes user $u$'s  number of sessions. Each session $s_{u,i}$ comprises a set of item-timestamp tuples $\{(v_{i,1}, t_{i,1}), \dots, (v_{i,n_i}, t_{i,n_i})\}$ where $n_i$ denotes the number of items in the $i$-th session of user $u$.
\end{definition}

\begin{figure}[t]
    \centering
    {\includegraphics[width=0.95\columnwidth]{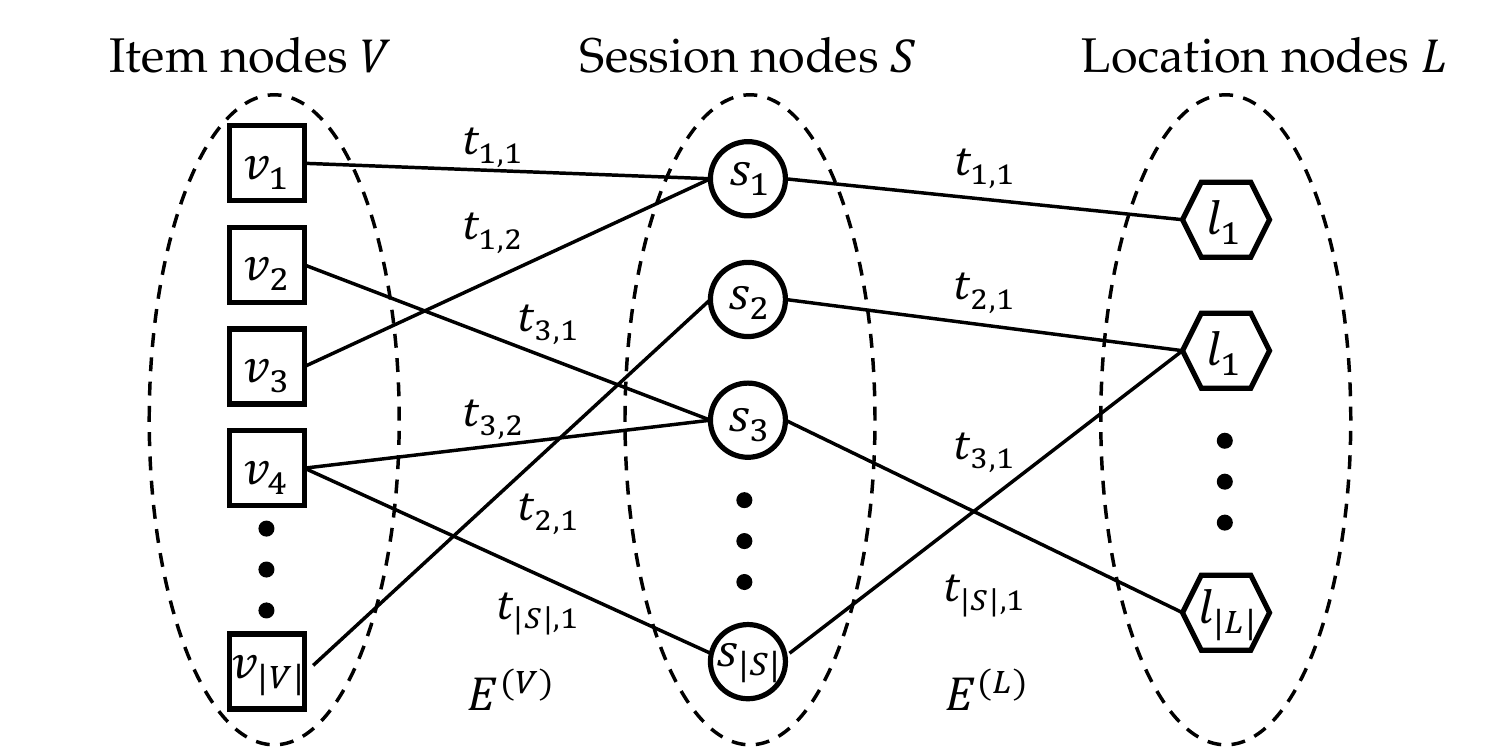}}
    \vspace{-0.1in}
    \caption{Schematic view of user spatiotemporal behavior graph $G_u$. This tripartite graph consists of user's sessions $S$, locations $L$, and items $V$ as nodes; and, $E^{(V)}$ of session-item edges and $E^{(L)}$ of session-location edges.}
    \vspace{-0.15in}
    \label{fig:user_graph}
\end{figure}

In a large-scale spatiotemporal behavior log, each user $u \in \mathcal{U}$ is associated with a subset of sessions $S_u \subseteq \mathcal{S}$, a subset of items $V_u \subseteq \mathcal{V}$ have been interacted with, and a subset of locations $L_u \subseteq \mathcal{L}$. Each session $s_{u,i} \in S_u$ is paired with a geographical location signal $l_{u,i} \in L_u$ and each item $v_{i,j} \in V_u$ is paired with an explicit timestamp $t_{i,j}$ forming a behavior entry. To capture the complex temporal and spatial patterns in the spatiotemporal behavior log, we represent a user's behaviors as a tripartite graph structure $G_u$ as shown in Figure \ref{fig:user_graph}. The graph $G_u$ is defined on $S_u$, $L_u$ and $V_u$, along with the their corresponding relationships. (Without causing ambiguity, we reduce the subscript $u$ on $S_u$, $L_u$ and $V_u$ for brevity.)

\begin{definition}[User Spatiotemporal Behavior Graph]
\label{def:user_graph}
A user $u$'s spatiotemporal behavior graph $G_u=(S, L, V, E)$ includes the user's sessions $S$, locations $L$ and items $V$ as nodes. There exists an edge $(s_{i}, l_{i}) \in E^{(L)} \subseteq E$ between a session node $s_{i} \in S$ and a location node $l_{i} \in L$ if the user started the session at this location. And, there exists an edge $(s_i, v_{i,j}) \in E^{(V)} \subseteq E$ between a session node $s_{i} \in S$ and an item node $v_{i,j} \in V$ if the user interacted with this item within the session. Each edge of $E$ possesses a time attribute indicating the temporal signal of the interaction between two nodes.
\end{definition}

The pairing timestamp $t_{i,j}$ ($i < m_u$, $j < n_{i}$) for each item in the behavior log can be directly used as the time attribute value for any edge of $E^{(V)}$. For an edge between a session node and a location node of $E^{(L)}$, we use the timestamp of the first item in the session, i.e., the leading timestamp $t_{i,1}$ of the session, as the time attribute value. Note that the subset of edges $E^{(V)}$ describe the many-to-many relationships between the session nodes $S$ and item nodes $V$, whereas the subset of edges $E^{(L)}$ describe the one-to-many relationships between location nodes $L$ and session nodes $S$. By modeling each user's behaviors as a spatiotemporal behavior graph $G$, we are able to format the spatiotemporal behavior log as:

\begin{definition}[Spatiotemporal Behavior Graph Data]
\label{def:spatiotemporal_data}
A spatiotemporal behavior graph data $\mathcal{B}=(\mathcal{G}, \mathcal{A})$ represent each user $u$ as a user spatiotemporal behavior graph $G_u=(S, L, V, E) \in \mathcal{G}$, and is related to a specific label $a_u \in \mathcal{A}$ where $\mathcal{A}$ can be categorical or numerical. All user spatiotemporal behavior graphs $\forall G_u \in \mathcal{G}$ share the same sets of sessions $\mathcal{S}$, items $\mathcal{V}$ and locations $\mathcal{L}$.
\end{definition}

After we have formatted the spatiotemporal behavior graph data, we can now formally define our research problem as:

\noindent \textbf{Problem:} 
\textbf{Given} a spatiotemporal behavior graph data $\mathcal{B}=(\mathcal{G}, \mathcal{A})$ on a set of users $\mathcal{U}$, \textbf{learn} an embedding function $f$ that can map each user $u \in \mathcal{U}$, denoted by her spatiotemporal behavior graph $G_u \in \mathcal{G}$, in to a low-dimensional hidden representation $\mathbf{u}$, i.e., $f: \mathcal{G} \mapsto \mathbb{R}^{K_{\mathcal{U}}}$, where $K_{\mathcal{U}}$ is the dimensionality of vector $\mathbf{u}$ ($K_{\mathcal{U}} << |\mathcal{U}|, |\mathcal{S}|, |\mathcal{V}|, |\mathcal{L}|$). The user embedding vector $\mathbf{u}$ should (1) capture the spatial pattern and temporal patterns of different periodicity in the user's behaviors, and (2) be highly indicative about the corresponding label $a_u \in \mathcal{A}$.
\vspace{0.08in}

%% file: 4approach.tex
\begin{figure*}[t]
	\centering
	\vspace{-0.1in}
	{\includegraphics[width=0.83\textwidth]{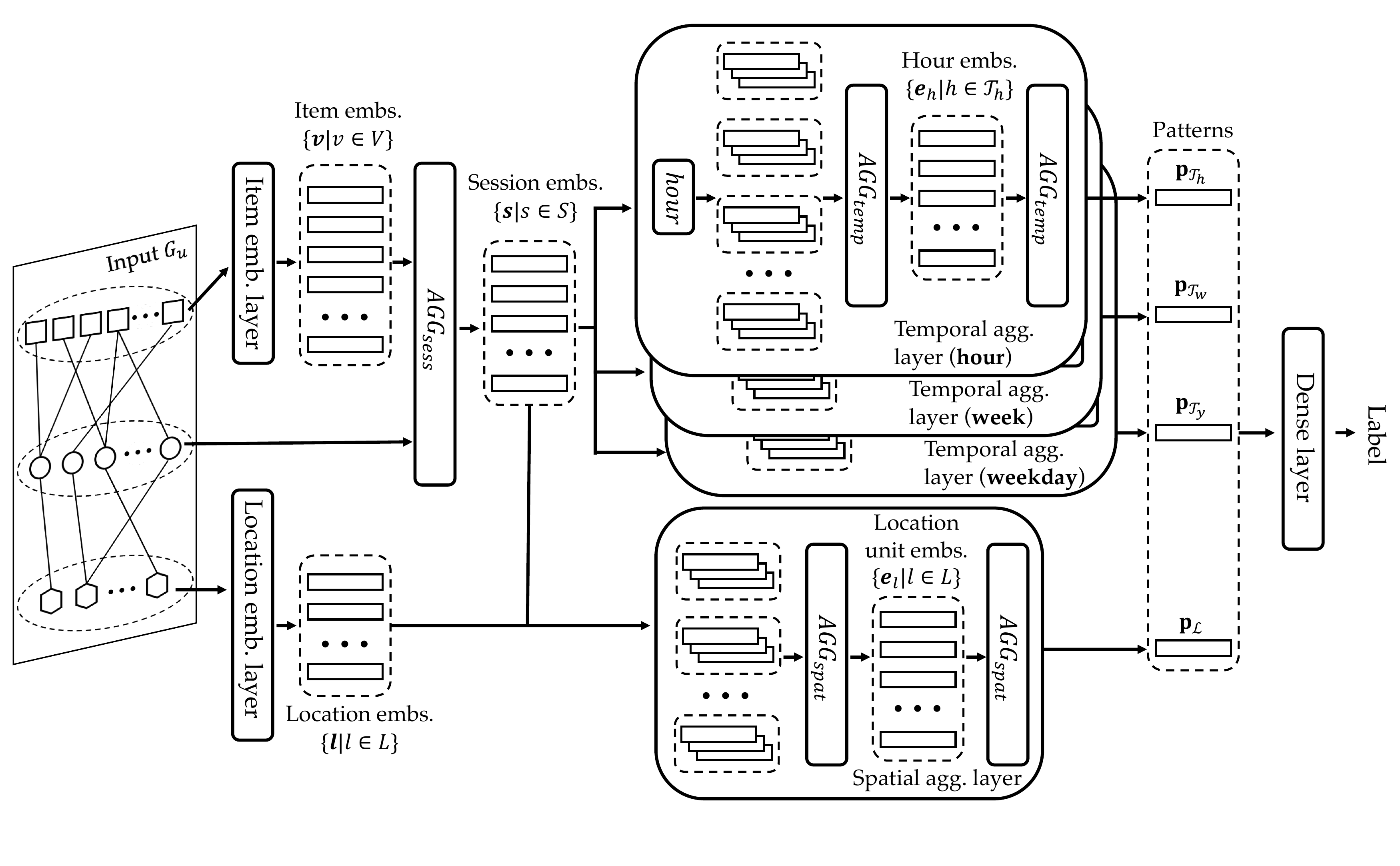}}
	\vspace{-0.35in}
	\caption{{CalendarGNN} architecture: Session embeddings are generated by aggregating its item embeddings. The embeddings of sessions are aggregated into hour, week, weekday unit embeddings, and location unit embeddings. Next, embeddings of temporal/spatial units are aggregated into pattern embeddings, and further fused into the user embedding for prediction.}
	\vspace{-0.15in}
	\label{fig:calendargnn}
\end{figure*}

In this section, we present a novel deep architecture \textsc{CalendarGNN} for predicting user attributes by learning user's spatiotemporal behavior patterns.
The overall design is shown in Figure \ref{fig:calendargnn}.
We first introduce the item and location embedding layers for embedding the heterogeneous features of item and location nodes in the input user spatiotemporal behavior graph into initial embeddings;
then, we present the spatiotemporal aggregation layers as core functions for generating spatial and temporal unit embeddings;
next, we describe the aggregation and fusion of different spatial and temporal patterns as user representation, and the subsequent predictive model.
At last, to capture the interactions between the spatial pattern and various temporal patterns, we present an enhanced model variant \textsc{CalendarGNN-Attn} that employs an interactive attention mechanism to dynamically adapt importances of different patterns.

\subsection{Item and Location Embedding Layers}

The inputs into \textit{CalendarGNN} are a user spatiotemporal behavior graphs $G_u=(S, L, V, E)$ and all users $\forall u \in \mathcal{U}$ share the same space of items $\bigcup V =\mathcal{V}$ and locations $\bigcup L=\mathcal{L}$. The first step of \textit{CalendarGNN} is to embed all items $\mathcal{V}$ and locations $\mathcal{L}$ of heterogenous features into their initial embeddings. Figure \ref{fig:emb_layers} illustrates the design of the item embedding layer and the location embedding layer.

\subsubsection{Item embedding layer}
An item $v \in \mathcal{V}$ such as a news article can be described by a group of heterogeneous features: (i) the identification, e.g., the ID of article; (ii) the topic, e.g., the category of article; and, (iii) the content, e.g., the title of the article. For each item, we feed its raw features into the item embedding layer (shown in Figure \ref{fig:item_emb_layer}) to generate the initial embedding. Particularly, for categorical features such as the item ID and category, we use \textit{Multilayer Perceptron} (\textsc{MLP}) to embed them into dense hidden representations; and, for textual feature, i.e., the item title, we use \textit{Bidirectional Long Short-Term Memory} (\textsc{BiLSTM}) \cite{schuster1997bidirectional} encoder to generate its hidden representation. Then, the embeddings of different features are concatenated together as the item embedding $\mathbf{v} \in \mathbb{R}^{K_{\mathcal{V}}}$ where $K_{\mathcal{V}}$ is the dimensions of the item embedding vector.

\subsubsection{Location embedding layer} 
\begin{figure}[t]
    \centering
    \subfigure[Item embedding layer]
    {\includegraphics[width=0.475\columnwidth]{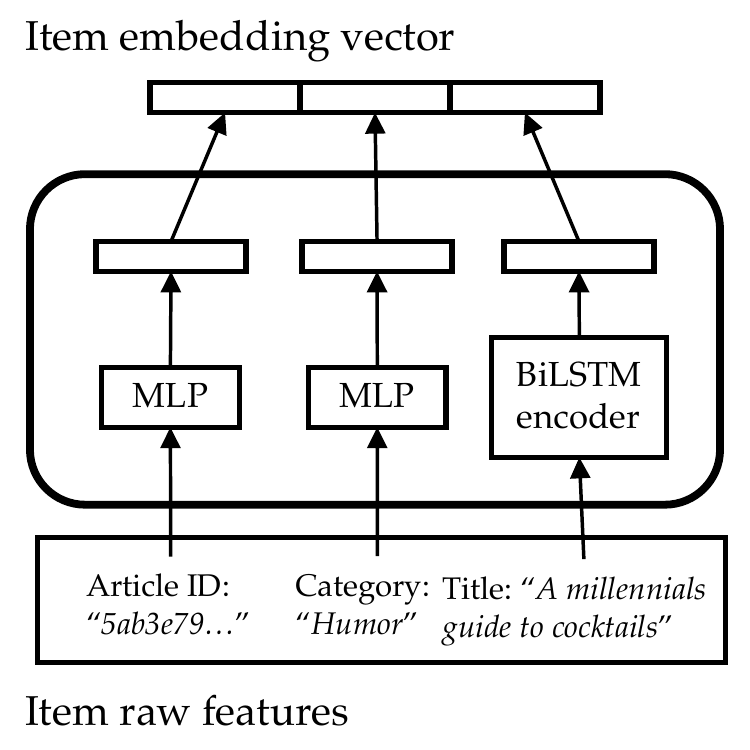} \label{fig:item_emb_layer}}
    \subfigure[Location embedding layer]
    {\includegraphics[width=0.475\columnwidth]{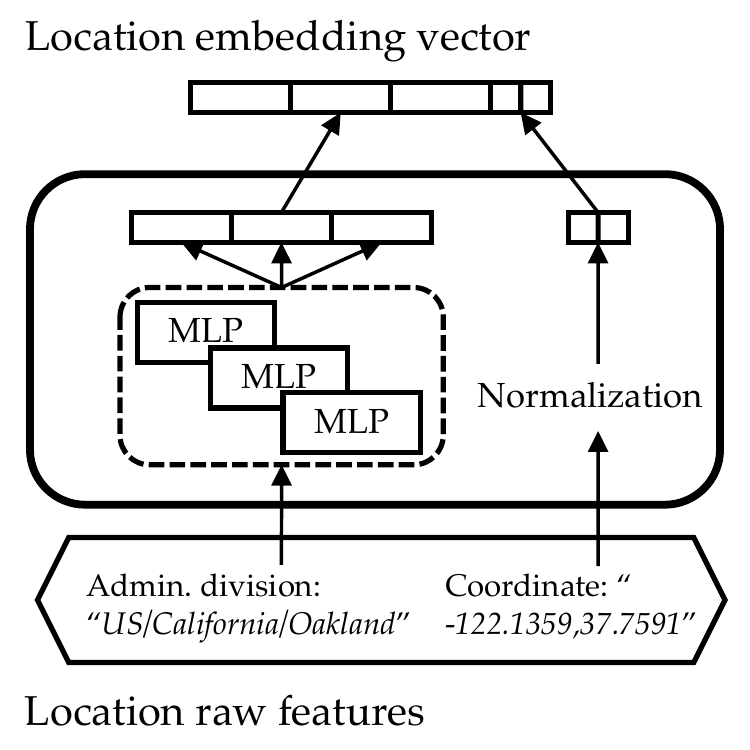} \label{fig:loc_emb_layer}}
    \vspace{-0.2in}
    \caption{The item embedding layer (left) takes raw features of an item, i.e., the ID, category and title, as input and generates its embedding vector; and, the location embedding layer (right) takes the administrative division and coordinate of a location as input and generates its embedding vector.}
    \label{fig:emb_layers}
    \vspace{-0.25in}
\end{figure}

Each location $l \in \mathcal{L}$ is denoted by a multi-level administrative division name in the format of ``\textit{county/region/city}'', and a coordinate point of longitude and latitude. One example location is ``\textit{US/California/Oakland}'' and its coordinate ``\textit{-122.1359, 37.7591}''. We use three distinct \textsc{MLPs} to encode the administrative division at different levels which could be partially empty. The outputs are concatenated with normalized coordinates (shown in Figure \ref{fig:loc_emb_layer}) as the location embedding vector $\mathbf{l} \in \mathbb{R}^{K_{\mathcal{E}}}$.

\subsection{Spatiotemporal Aggregation Layer}

After item and location nodes are embedded into initial embeddings,
\textsc{CalendarGNN} generates the embeddings of session nodes by aggregating from item embeddings. For a session node $s_i \in S$ in $G_u=(S, L, V, E)$, its embedding vector $\mathbf{s}_i$ is generated by applying an aggregation function $\textsc{Agg}_{\text{sess}}$ on all item nodes linked to it:
\begin{equation}
	\label{eqn:session_agg}
    \mathbf{s}_i = \sigma \left( \mathbf{W}_{\mathcal{S}} \cdot \textsc{Agg}_{\text{sess}} ( \{\mathbf{v}_{i,j}~|~\forall{(s_i, v_{i,j})} \in E\} ) + \mathbf{b}_{\mathcal{S}} \right),
\end{equation}
where $\sigma$ is a function for non-linearity, such as \textsc{ReLU} \cite{nair2010rectified}; and, $\mathbf{W}_{\mathcal{S}}$ and $\mathbf{b}_{\mathcal{S}}$ are parameters to be learned. The weight matrix $\mathbf{W}_{\mathcal{S}} \in \mathbb{R}^{K_\mathcal{S} \times K_\mathcal{V} }$ transforms the $K_\mathcal{V}$-dim item embedding space to the $K_\mathcal{S}$-dim session embedding space (assuming $\textsc{Agg}_{\text{sess}}$ has the same number of input and output dimensions). The aggregation function $\textsc{Agg}_{\text{sess}}$ can be arbitrary injective function for mapping a set of vectors into an output vector. Since the session node's neighbor of item nodes $\{{v}_{i,j}~|~\forall{(s_i, v_{i,j})} \in E\}$ can naturally be ordered by their timestamps $t_{i,j}$, we arrange items as sequence and choose to use \textit{Gated Recurrent Unit} (\textsc{GRU}) \cite{cho2014learning} as the $\textsc{Agg}_{\text{sess}}$ function.

Now, we have generated session node embeddings $\{\mathbf{s}~|~s\in S \}$ for $G_u$, \textsc{CalendarGNN} is ready to generate spatial and temporal patterns.
The core intuition is to inject external knowledge about the calendar system's structure into the architecture of \textsc{CalendarGNN} so that we can aggregate a user's session node embeddings into spatial pattern and temporal patterns of various periodicity based on their spatial and temporal signals.
Specifically, we pass session node embeddings to: (1) the temporal aggregation layer for generating temporal patterns of various periodicity; and, (2) the spatial aggregation layer for generating spatial pattern.

\subsubsection{Temporal aggregation layer}
Given session node embeddings $\{\mathbf{s}~|~s\in S \}$ of $G_u$, the idea of temporal aggregations in this layer is to: (1) map sessions $S$'s continuous timestamps into a set of discrete time units, and (2) aggregate sessions of the same time unit into the corresponding time unit embeddings, and, (3) aggregate time unit embeddings into the embedding of temporal pattern.

Mapping sessions $S$' timestamps $\{t_i~|~s_i \in S\}$ into set of discrete time units is analogous to bucket session embeddings by discrete time units. We regard the leading timestamp of corresponding item nodes as the session's timestamp, i.e., $t_i=\text{min}(\{t_{i,j}~|~\forall{(s_i, v_{i,j})} \in E\})$.
Particularly, taken inspiration from the daily calendar system, we convert $t_{i}$ into three types of time units:
\begin{compactitem}
    \item $h_{i} = hour(t_{i}) \in \mathcal{T}_{h}$, where $\mathcal{T}_{h}$ has 24 distinct values: 0AM, 1AM, ..., 11PM;
    \item $w_{i} = week(t_{i}) \in \mathcal{T}_{w}$, where $\mathcal{T}_{w}$ is the set of weeks of the year, e.g., Week 18;
    \item $y_{i} = weekday(t_{i}) \in \mathcal{T}_{y}$, where $\mathcal{T}_{y}$ has 7 values: Sunday, Monday, ..., Saturday.
\end{compactitem}

The time unit mapping functions $hour$, $week$ and $weekday$ takes a timestamp as input and outputs a specific time unit.
The cardinality of the output time units set can vary, e.g., $|\mathcal{T}_{h}|=24$ or $|\mathcal{T}_{y}|=7$.
In this work, we leverage 3 time units of common sense, i.e., hour, week, and weekday, for capturing the complex time structures in user behaviors. \textsc{CalendarGNN} maintains the flexibility to model temporal pattern of arbitrary periodicity, such as daytime/night or minute, providing the new time unit mapping function(s).

Once the session nodes are mapped into specified time units, \textsc{CalendarGNN} aggregates the session node embeddings into various time unit embeddings by applying a temporal aggregation function $\textsc{Agg}_\text{temp}$ on sessions of the same time unit: 
\begin{eqnarray}
	\label{eqn:hour_unit_agg}
	\mathbf{e}_{h} = \sigma \left( \mathbf{W}_{h} \cdot \textsc{Agg}_\text{temp} (\{\mathbf{s}_{i}~|~h_{i} = h \in \mathcal{T}_{h} \}) + \mathbf{b}_{h} \right), \\
	\label{eqn:week_unit_agg}
	\mathbf{e}_{w} = \sigma \left( \mathbf{W}_{w} \cdot \textsc{Agg}_\text{temp} (\{\mathbf{s}_{i}~|~w_{i} = w \in \mathcal{T}_{w} \}) + \mathbf{b}_{w} \right), \\
	\label{eqn:weekday_unit_agg}
	\mathbf{e}_{y} = \sigma \left( \mathbf{W}_{y} \cdot \textsc{Agg}_\text{temp} (\{\mathbf{s}_{i}~|~y_{i} = y \in \mathcal{T}_{y} \}) + \mathbf{b}_{y} \right),
\end{eqnarray}
\noindent where the weight matrices $\mathbf{W}_{h} \in \mathbb{R}^{K_{h} \times K_\mathcal{S}}$, $\mathbf{W}_{w} \in \mathbb{R}^{K_{w} \times K_\mathcal{S}}$ and $\mathbf{W}_{y} \in \mathbb{R}^{K_{y} \times K_\mathcal{S}}$ transform the $K_{\mathcal{S}}$-dim session embedding space into $K_{h}$-dim hour embedding space, $K_{w}$-dim week embedding space, and $K_{y}$-dim weekday embedding space, respectively. The choice of $\textsc{Agg}_\text{temp}$ is also set to \textsc{GRU} since all items of the same time unit can naturally be ordered by their raw timestamp.

Next, these time unit embeddings in the three dimensions (i.e., hour, week, and weekday) are further aggregated into embeddings of respective temporal patterns:
\begin{eqnarray}
	\label{eqn:hour_pattern_agg}
	\mathbf{p}_{\mathcal{T}_{h}} = \sigma \left( \mathbf{W}_{\mathcal{T}_{h}} \cdot \textsc{Agg}_\text{temp} (\{\mathbf{e}_{h}~|~\forall{h} \in \mathcal{T}_{h} \}) + \mathbf{b}_{\mathcal{T}_{h}} \right), \\
	\label{eqn:week_pattern_agg}
	\mathbf{p}_{\mathcal{T}_{w}} = \sigma \left( \mathbf{W}_{\mathcal{T}_{w}} \cdot \textsc{Agg}_\text{temp} (\{\mathbf{e}_{w}~|~\forall{w} \in \mathcal{T}_{w} \}) + \mathbf{b}_{\mathcal{T}_{w}} \right), \\
	\label{eqn:weekday_pattern_agg}
	\mathbf{p}_{\mathcal{T}_{y}} = \sigma \left( \mathbf{W}_{\mathcal{T}_{y}} \cdot \textsc{Agg}_\text{temp} (\{\mathbf{e}_{y}~|~\forall{y} \in \mathcal{T}_{y} \}) + \mathbf{b}_{\mathcal{T}_{y}} \right),
\end{eqnarray}
where the weight matrices $\mathbf{W}_{\mathcal{T}_{h}} \in \mathbb{R}^{K_{\mathcal{T}_{h}} \times {K_{h}}}$, $\mathbf{W}_{\mathcal{T}_{w}} \in \mathbb{R}^{K_{\mathcal{T}_{w}} \times {K_{w}}}$, $\mathbf{W}_{\mathcal{T}_{y}} \in \mathbb{R}^{K_{\mathcal{T}_{y}} \times {K_{y}}}$ transform the aggregated hour, week, and weekday embeddings into the corresponding ($K_{\mathcal{T}_{h}}$-dim) hourly, ($K_{\mathcal{T}_{w}}$-dim) weekly, and ($K_{\mathcal{T}_{h}y}$-dim) weekday patterns, respectively. Each one of these temporal pattern captures the user's temporal behavior pattern of a specific periodicity.

In addition to temporal patterns, another indispensable aspect of user's behavior pattern relates to the spatial signals of sessions.
\textsc{CalendarGNN} is capable of discovering user's spatial pattern by aggregating session embeddings via the spatial aggregation layer.

\subsubsection{Spatial aggregation layer} Similar to the treatment of temporal aggregation layer previous introduced, for generating spatial pattern, \textsc{CalendarGNN} first aggregates the session node embeddings into location unit embeddings based on their spatial signals:
\begin{eqnarray}
	\label{eqn:spat_unit_agg}
	\mathbf{e}_{l} = \sigma \left( \mathbf{W}_{\mathcal{S}\times\mathcal{L}} \cdot \textsc{AGG}_\text{spat} (\{\mathbf{s}_{i} \oplus \mathbf{l}_{i}~|~l_{i} = l \in {L} \} + \mathbf{b}_{\mathcal{S}\times\mathcal{L}} \right),
\end{eqnarray}
\noindent where $\oplus$ is concatenation operator, and $\mathbf{W}_{\mathcal{S}\times\mathcal{L}} \in \mathbb{R}^{K_{l} \times (K_\mathcal{S}+K_\mathcal{E})}$ transforms the concatenated space of session embedding initial location embedding into the location unit embedding space, and $\textsc{AGG}_\text{spat}$ is the spatial aggregation function. We also arrange sessions of the same location unit by their timestamps and choose to use \textsc{GRU} as $\textsc{AGG}_\text{spat}$.

Then, \textsc{CalendarGNN} aggregates various location unit embeddings into the embedding vector of spatial pattern:
\begin{eqnarray}
	\label{eqn:spat_pattern_agg}
	\mathbf{p}_{\mathcal{L}} = \sigma \left( \mathbf{W}_{\mathcal{L}} \cdot \textsc{AGG}_\text{spat} (\{\mathbf{e}_{l}~|~\forall{l} \in {L}\}) + \mathbf{b}_{\mathcal{L}} \right),
\end{eqnarray}
\noindent where $\mathbf{W}_{\mathcal{L}} \in \mathbb{R}^{K_\mathcal{L} \times {K_{l}}}$ transforms the location unit embedding space into the spatial pattern space.

By feeding the session node embeddings into temporal aggregation layers and spatial aggregation layer, \textsc{CalendarGNN} has generated temporal patterns. i.e., $\mathbf{p}_{\mathcal{T}_{h}}$, $\mathbf{p}_{\mathcal{T}_{w}}$ and $\mathbf{p}_{\mathcal{T}_{y}}$, and the spatial pattern, i.e., $\mathbf{p}_{\mathcal{L}}$.
At last, \textsc{CalendarGNN} fuses all temporal patterns and spatial pattern into a holistic user latent representation $\mathbf{u}$, and pass it to the subsequent predictive model for prediction and output.

\subsection{Fusion of Patterns and Prediction}
\label{subsec:fusion}

To get the latent representation of user, we concatenate all temporal patterns and the spatial pattern together:
\begin{equation}
	\label{eqn:user_emb_concat}
    \mathbf{u} = \mathbf{p}_{\mathcal{T}_{h}} \oplus \mathbf{p}_{\mathcal{T}_{w}} \oplus \mathbf{p}_{\mathcal{T}_{y}} \oplus \mathbf{p}_{\mathcal{L}} \in \mathbb{R}^{K_\mathcal{U}},
\end{equation}
where ${K_\mathcal{U}}={K_{\mathcal{T}_{h}}}+{K_{\mathcal{T}_{w}}}+{K_{\mathcal{T}_{y}}}+{K_{\mathcal{L}}}$.

We use a single dense layer as the final predictive model for generating user attribute predictions. The discrepancy between the output of the last dense layer and the target attribute value is measured by the objective function for optimization. Specifically, if the user label $a_u \in \mathcal{A}$ is a categorical value, i.e., the task is multi-class classification (with binary classification as a special case), we employ the following cross-entropy objective function:
\begin{equation}
	\label{eqn:loss_classification}
    \mathcal{J} = - \sum\limits_{u \in \mathcal{U}} \sum\limits_{a \in \mathcal{A}} \mathbb{I}_{a_u=a} \cdot \frac{\exp \left({\mathbf{W}_a \cdot \mathbf{u}}\right)}{\sum\limits_{a' \in \mathcal{A}}{\exp \left({\mathbf{W}_a \cdot \mathbf{u}}\right)}},
\end{equation}
where $\mathbf{W}_a \in \mathbb{R}^{K_\mathcal{U}}$ is the weight vector for label $a \in \mathcal{A}$ and $\mathbb{I}$ is an indicator function. If the label is a numerical value ($a_u \in \mathbb{R}$), we employ the following objective function for the regression task:
\begin{equation}
	\label{eqn:loss_regression}
    \mathcal{J} = - \sum\limits_{u \in \mathcal{U}} {\left( \mathbf{W} \cdot \mathbf{u} - a_u \right)}^2.
\end{equation}

\subsection{Interactive Spatiotemporal Patterns}
\label{sec:interactive}
By utilizing the temporal and spatial aggregation layers, \textsc{CalendarGNN} is able to generate spatial pattern and temporal patterns of different periodicity (Eqn. (\ref{eqn:hour_pattern_agg}) to (\ref{eqn:spat_pattern_agg})).
However, there are a few limitations. 
First, different temporal/spatial unit embeddings are of different importance levels to its pattern and this should be reflected during the pattern generation process. 
Secondly, there could be rich interactions between the spatial pattern and different temporal patterns. These interactions should be carefully captured by the model and be reflected in the true spatiotemporal patterns \cite{he2017neural, wu2018restful}.

To address these limitations, we propose a model variant that employs an interactive attention mechanism \cite{ma2017interactive} and denote it as \textsc{CalendarGNN-Attn}. It enables interactions between spatial and temporal patterns by summarizing location unit embeddings and a certain type of time unit embeddings into an interactive spatiotemporal pattern. For location unit embeddings $\{\mathbf{e}_{l}~|~\forall{l} \in {L}\}$ and time unit embeddings such as hour embeddings $\{\mathbf{e}_{h}~|~\forall{h} \in \mathcal{T}_{h} \}$, a location query and a temporal query are first generated:
\begin{eqnarray}
	\label{eqn:attention_query}
    \bar{\mathbf{e}}_{l} = \sum_{{l} \in {L}} \mathbf{e}_{l}/ \vert \mathcal{L} \vert, 
    \bar{\mathbf{e}}_{h} = \sum_{{h} \in \mathcal{T}_{h}} \mathbf{e}_{h}/ \vert \mathcal{T}_{h} \vert,
\end{eqnarray}
where $\vert \cdot \vert$ denotes the cardinality of the set. On one hand, to consider the impacts from spatial signals on temporal signals, a attention weight vector $\mathbf{\alpha}_{h}^{(\mathcal{L},\mathcal{T}_{h})}$ is generated using the location query vector $\bar{\mathbf{e}}_{l}$ and the temporal unit embeddings $\{\mathbf{e}_{h}~|~\forall{h} \in \mathcal{T}_{h} \}$:
\begin{equation}
	\label{eqn:attn_w_spat_hour}
    \mathbf{\alpha}_{h}^{(\mathcal{L},\mathcal{T}_{h})} = \frac{\exp \left( f(\mathbf{e}_{h}, \bar{\mathbf{e}}_{l})\right)}{\sum_{{h} \in \mathcal{T}_{h}} \exp \left( f(\mathbf{e}_{h}, \bar{\mathbf{e}}_{l})\right)},
\end{equation}
where $f$ is a function for scoring the importance of $\mathbf{e}_{h}$ w.r.t. the location query $\mathbf{e}_{l}$ and is defined as:
\begin{equation}
	\label{eqn:attn_bilinear}
    f(\mathbf{e}_{h}, \bar{\mathbf{e}}_{l}) = \tanh \left(\mathbf{e}_{h} \cdot \mathbf{W}_{(\mathcal{L},\mathcal{T}_{h})} \cdot \bar{\mathbf{e}}_{l}^{T} + \mathbf{b}_{(\mathcal{L},\mathcal{T}_{h})}  \right),
\end{equation}
where $\mathbf{W}_{(\mathcal{L},\mathcal{T}_{h})}$ is the weight matrix of a bilinear transformation. Thus, we are able to generate the temporal pattern under impacts from the location units as:
\begin{equation}
	\label{eqn:attn_hour_pattern_spat}
    \mathbf{p}_{\mathcal{T}_{h}}^{\mathcal{L}} = \sum_{{h} \in \mathcal{T}_{h}} \mathbf{\alpha}_{h}^{(\mathcal{L},\mathcal{T}_{h})} \mathbf{e}_{h}.
\end{equation}

On the other hand, we also consider the impacts from temporal signals on locations signals. So the attention weight vector for location unit embeddings can be calculated as:
 \begin{equation}
	\label{eqn:attn_w_hour_spat}
    \mathbf{\alpha}_{l}^{(\mathcal{T}_{h}, \mathcal{L})} = \frac{\exp \left( f(\mathbf{e}_{l}, \bar{\mathbf{e}}_{h})\right)}{\sum_{{l} \in {L}} \exp \left( f(\mathbf{e}_{l}, \bar{\mathbf{e}}_{h})\right)},
\end{equation}
and the spatial pattern under impacts from the time units is:
\begin{equation}
	\label{eqn:attn_hour_pattern_spat}
    \mathbf{p}_{\mathcal{L}}^{\mathcal{T}_{h}} = \sum_{{l} \in {L}} \mathbf{\alpha}_{l}^{(\mathcal{T}_{h}, \mathcal{L})} \mathbf{e}_{l}.
\end{equation}

Then, these two one-way impacted spatiotemporal patterns are concatenated to get the interactive spatiotemporal pattern:
\begin{equation}
	\label{eqn:attn_spat_hour_pattern}
    \mathbf{p}_{\mathcal{L}, \mathcal{T}_{h}} = \mathbf{p}_{\mathcal{T}_{h}}^{\mathcal{L}} \oplus \mathbf{p}_{\mathcal{L}}^{\mathcal{T}_{h}}.
\end{equation}

Similarly, we can generate the interactive spatiotemporal patterns for the other two type of time units of week $\mathbf{p}_{\mathcal{L}, \mathcal{T}_{w}}$ and weekday $\mathbf{p}_{\mathcal{L}, \mathcal{T}_{y}}$. Then, the final user representation is:
\begin{equation}
	\label{eqn:user_emb_interactive}
    \mathbf{u} = \mathbf{p}_{\mathcal{L}, \mathcal{T}_{h}} \oplus \mathbf{p}_{\mathcal{L}, \mathcal{T}_{w}} \oplus \mathbf{p}_{\mathcal{L}, \mathcal{T}_{y}}.
\end{equation}
Thus, by substituting Eqn. (\ref{eqn:user_emb_interactive}) into Eqn. (\ref{eqn:user_emb_concat}), \textsc{CalendarGNN-Attn} considers all interactions between the spatial pattern and temporal patterns when making predictions of user attributes.

%% file: 5experiments.tex

In this section, we evaluate the proposed model on 2 real-world spatiotemporal behavior datasets. The empirical analysis covers: (1) effectiveness, (2) explainability, and (3) robustness and efficiency.

\begin{table}[t]
	\renewcommand{\arraystretch}{1.15}
	\centering
	\caption{Summary statistics on two real-world spatiotemporal datasets $\mathcal{B}^{(w1)}$ and $\mathcal{B}^{(w2)}$.}
	\label{tab:datasets}
	\vspace{-0.1in}
	\Scale[1]{
	\begin{tabular}{|l||c|c|c|c|c|}
	\hline
		\textbf{Dataset} & $|\mathcal{U}|$ & $|\mathcal{V}|$ & $|\mathcal{L}|$ & $|\mathcal{S}|$ & Avg. $|G|$ \\ \hline \hline
		$\mathcal{B}^{(w1)}$ & {10,545} & {7,984} & {7,393} & {651,356} & {242.8} \\ \hline
		$\mathcal{B}^{(w2)}$ & {8,017} & {6,389} & {4,445} & {135,805} & {61.3} \\ \hline
	\end{tabular}
	}
	\vspace{-0.15in}
\end{table}

\subsection{Datasets}
We collected large-scale user behavior logs from 2 real portal websites providing news updates and articles on various topics, and created 2 spatiotemporal datasets $\mathcal{B}^{(w1)}$ and $\mathcal{B}^{(w2)}$. They contain users' spatiotemporal behavior log of browsing these 2 websites and both datasets range from Jan. 1 2018 to Jun. 30 2018. After all users have been anonymized, we filtered each dataset to keep around $10,000$ users with most clicks. More statistics are provided in Table \ref{tab:datasets}. The 3 user attributes used for prediction tasks are:
\begin{compactitem}
	\item {$\mathcal{A}^{(gen)}$}: the binary gender of user $\forall a^{(gen)} \in \{\text{``f''}, \text{``m''}\}$ where ``f'' denotes female and ``m'' denotes male,
	\item {$\mathcal{A}^{(inc)}$}: the categorical income level of user such that $\forall a^{(inc)} \in \{0, 1, \dots, 9\}$ where larger value indicate higher annual household income level and 0 indicates unknown,
	\item {$\mathcal{A}^{(age)}$}: the calculated age of user based on registered birthday. This label is treated as real value in all experiments.
\end{compactitem}

\begin{table*}[t]
	\renewcommand{\arraystretch}{1.1}
	\centering
	\caption{For dataset $\mathcal{B}^{(w1)}$, the performance of \textsc{CalendarGNN},  \textsc{CalendarGNN-Attn} (\textsc{CalGNN-Attn}), and baseline methods on predicting user attributes. For all metrics except error-based MAE and RMSE, higher values indicate better performance.}
	\label{tab:results_nyr}
	\vspace{-0.15in}
	\Scale[0.925]{
	\begin{tabular}{|l||c|c|c|c|c|c|c|c|c|c|c|c|}
	\hline
		\multirow{2}*{\textbf{Method}} & \multicolumn{4}{|c|}{\textbf{Gender} $\mathcal{A}^{(gen)}$} & \multicolumn{4}{|c|}{\textbf{Income} $\mathcal{A}^{(inc)}$} & \multicolumn{4}{|c|}{\textbf{Age} $\mathcal{A}^{(age)}$} \\ \cline{2-13}
		 & {$Acc.$} & {$AUC$} & {$F1$} & {$MCC$} & {$Acc.$} & {$F1$-macro} & {$F1$-micro} & {Cohen's kappa $\kappa$} & {$R^{2}$} & {$MAE$} & {$RMSE$} & {Pearson's $r$}  \\ \hline \hline
		\textsc{LR}					& {67.08\%} & {.6469} & {.6628} & {.3319} & {19.54\%} & {.0642} & {.1957} & {.0121} & {.0349} & {12.22} & {15.53} & {.2938} \\ \hline
		\textsc{LearnSuc}			& {67.41\%} & {.6541} & {.6680} & {.3330} & {14.58\%} & {.0531} & {.1523} & {.0078} & {.0523} & {12.18} & {15.49} & {.2989}  \\ \hline
		\textsc{SR-GNN} 				& {69.82\%} & {.6733} & {.6854} & {.3510} & {20.21\%} & {.0676} & {.1949} & {.0182} & {.0121} & {15.20} & {16.87} & {.2566} \\ \hline
		\hline
		\textsc{ECC}		 		& {70.29\%} & {.6886} & {.6832} & {.3825} & {23.54\%} & {.0767} & {.2267} & {.0222} & {.2158} & {11.12} & {13.88} & {.4768}  \\ \hline
		\textsc{DiffPool} 			& {72.12\%} & {.7189} & {.7089} & \textbf{.4514} & {25.87\%} & {.0928} & {.2763} & {.0760} & {.2398} & \textbf{10.55} & {13.81} & {.4992}  \\ \hline
		\textsc{DGCNN}		 		& {71.26\%} & {.7129} & {.7068} & {.4189} & {24.55\%} & {.0879} & {.2509} & {.0687} & {.2351} & {10.86} & {13.97} & {.4809}  \\ \hline
		\textsc{CapsGNN}		 		& {70.85\%} & {.6979} & {.6921} & {.4031} & {23.71\%} & {.0750} & {.2189} & {.0378} & {.2270} & {10.90} & {13.86} & {.4645}  \\ \hline
		\textsc{SAGPool}		 		& {71.95\%} & {.7156} & {.7093} & {.4467} & {26.13\%} & {.0942} & {.2554} & {.0797} & {.2350} & {10.77} & {13.91} & {.4887}  \\ \hline
		\hline
		\textsc{CalendarGNN} 		& \textbf{72.98\%} & \textbf{.7250} & \textbf{.7119} & {.4503} & {28.83\%} & {.1059} & {.2981} & {.0887} & \textbf{.2412} & {10.57} & {13.60} & {.5033}  \\ \hline
		\textsc{CalGNN-Attn} 		& {72.70\%} & {.7236} & {.7112} & {.4491} & \textbf{29.67\%} & \textbf{.1100} & \textbf{.3062} & \textbf{.0910} & {.2401} & {10.65} & \textbf{13.52} & \textbf{.5069}  \\ \hline
	\end{tabular}
	}
	\vspace{-0.15in}
\end{table*}

\subsection{Experimental Settings}

\subsubsection{Baseline methods}
We compare \textsc{CalendarGNN} against state-of-the-art \textsc{GNN}-based methods:
\begin{compactitem}
	\item \textsc{ECC} \cite{simonovsky2017dynamic}: This method performs edge-conditioned convolutions over local graph neighborhoods and generate graph embedding with a graph coarsening algorithm.
	\item \textsc{DiffPool} \cite{ying2018hierarchical}: This method generates hierarchical representations of graph by learning a soft cluster assignment for nodes at each layer and iteratively merge nodes into clusters.
	\item \textsc{DGCNN} \cite{zhang2018end}: The core component \textsc{SortPooling} layer of this method takes unordered vertex features as input and outputs sorted graph representation vector of a fixed size.
	\item \textsc{CapsGNN} \cite{xinyi2018capsule}: This method extracts both node and graph embeddings as capsules and uses routing mechanism to generate high-level graph or class capsules for prediction.
	\item \textsc{SAGPool} \cite{lee2019self}: It uses self-attention mechanism on top of the graph convolution as a pooling layer and take the summation of outputs by each readout layer as embedding of the graph.
\end{compactitem}
Besides above GNN-based approaches, we also consider the following methods for modeling user behaviors in session-based scenario:
\begin{compactitem}
	\item {Logistic/Linear Regression (\textsc{LR})}: The former one is applied for classification tasks and the later one is used for regression task. The input matrix is a row-wise concatenation of user's item frequency matrix and location frequency matrix.
	\item \textsc{LearnSuc} \cite{wang2018multi}: This method considers user's sessions as behaviors denoted by multi-type itemset structure \cite{wang2019tube}. The embeddings of users, items, and locations are jointly learned by optimizing the collective success rate or the user label.
	\item \textsc{SR-GNN} \cite{wu2019session}: It uses graph structure to model user behavior of sessions and use \textsc{GNN} to generate node embeddings. The user session embedding is generated by concatenating the last item embedding and the aggregated items embedding.
\end{compactitem}

We use open-source implementations provided by the original paper for all baseline methods and follow the recommended setup guidelines when possible. 
Our code package is available on Github: \url{https://github.com/dmsquare/CalendarGNN}.

\subsubsection{Evaluation metrics.}
For classifying binary user label $\mathcal{A}^{(gen)}$,  we use metrics of \textit{mean accuracy} (Acc.), \textit{Area Under the precision-recall Curve} (AUC), F1 score and \textit{Matthews Correlation Coefficient} (MCC). For classifying multi-class user label $\mathcal{A}^{(inc)}$, metrics of \textit{mean accuracy} (Acc.), F1 (macro, micro) averaged score and \textit{Cohen's kappa} $\kappa$ are reported. For numerical user label $\mathcal{A}^{(age)}$, metrics of \textit{R-squared} ($R^{2}$), \textit{Mean Absolute Error} (MAE), \textit{Root-Mean-Square Error} (RMSE) and \textit{Pearson correlation coefficient} ($r$) are reported.

\subsection{Quantitative analysis}
Table \ref{tab:results_nyr} and \ref{tab:results_vnf} present the experimental results of \textsc{CalendarGNN} and baseline methods on classifying/predicting user labels $\mathcal{A}^{(gen)}$, $\mathcal{A}^{(inc)}$, and $\mathcal{A}^{(age)}$ on datasets $\mathcal{B}^{(w1)}$ and $\mathcal{B}^{(w2)}$, respectively.

\begin{table*}[t]
	\renewcommand{\arraystretch}{1.1}
	\centering
	\caption{For dataset $\mathcal{B}^{(w2)}$, the performance of \textsc{CalendarGNN},  \textsc{CalendarGNN-Attn} (\textsc{CalGNN-Attn}), and baseline methods on predicting user attributes. For all metrics except error-based MAE and RMSE, higher values indicate better performance.}
	\label{tab:results_vnf}
	\vspace{-0.15in}
	\Scale[0.925]{
	\begin{tabular}{|l||c|c|c|c|c|c|c|c|c|c|c|c|}
	\hline
		\multirow{2}*{\textbf{Method}} & \multicolumn{4}{|c|}{\textbf{Gender} $\mathcal{A}^{(gen)}$} & \multicolumn{4}{|c|}{\textbf{Income} $\mathcal{A}^{(inc)}$} & \multicolumn{4}{|c|}{\textbf{Age} $\mathcal{A}^{(age)}$} \\ \cline{2-13}
		 & {$Acc.$} & {$AUC$} & {$F1$} & {$MCC$} & {$Acc.$} & {$F1$-macro} & {$F1$-micro} & {Cohen's kappa $\kappa$} & {$R^{2}$} & {$MAE$} & {$RMSE$} & {Pearson's $r$}  \\ \hline \hline
		\textsc{LR}					& {66.53\%} & {.6410} & {.6523} & {.3100} & {18.21\%} & {.0655} & {.1887} & {.0097} & {.0320} & {12.79} & {16.92} & {.2763} \\ \hline
		\textsc{LearnSuc}			& {67.01\%} & {.6494} & {.6612} & {.3199} & {13.72\%} & {.0522} & {.1587} & {.0060} & {.0489} & {12.72} & {16.93} & {.2789}  \\ \hline
		\textsc{SR-GNN} 				& {67.80\%} & {.6562} & {.6660} & {.3289} & {19.79\%} & {.0686} & {.1910} & {.0201} & {.0209} & {15.88} & {17.08} & {.2370} \\ \hline
		\hline
		\textsc{ECC}		 		& {68.53\%} & {.6802} & {.6792} & {.3580} & {21.08\%} & {.0723} & {.2190} & {.0345} & {.2030} & {11.75} & {14.82} & {.4320}  \\ \hline
		\textsc{DiffPool} 			& {71.04\%} & {.6998} & {.6967} & {.4269} & {24.09\%} & {.0835} & {.2753} & {.0687} & {.2188} & {11.23} & {14.30} & {.4590}  \\ \hline
		\textsc{DGCNN}		 		& {70.20\%} & {.6972} & {.6855} & {.3892} & {22.70\%} & {.0809} & {.2472} & {.0600} & {.2180} & {11.49} & {14.69} & {.4392}  \\ \hline
		\textsc{CapsGNN}		 		& {68.29\%} & {.6806} & {.6800} & {.3588} & {21.92\%} & {.0789} & {.2196} & {.0438} & {.2059} & {11.82} & {14.69} & {.4389}  \\ \hline
		\textsc{SAGPool}		 		& {71.02\%} & {.7065} & {.6970} & {.4287} & {24.52\%} & {.0856} & {.2802} & {.0701} & {.2223} & {10.97} & {14.21} & {.4652}  \\ \hline
		\hline
		\textsc{CalendarGNN} 		& \textbf{71.63\%} & \textbf{.7104} & \textbf{.7038} & \textbf{.4389} & {27.10\%} & {.0909} & {.2798} & {.0742} & {.2223} & \textbf{10.79} & {13.88} & {.4872}  \\ \hline
		\textsc{CalGNN-Attn} 		& {71.47\%} & {.7098} & {.7021} & {.4341} & \textbf{28.17\%} & \textbf{.1015} & \textbf{.2964} & \textbf{.0846} & \textbf{.2332} & {10.87} & \textbf{13.67} & \textbf{.4963}  \\ \hline
	\end{tabular}
	}
	\vspace{-0.1in}
\end{table*}

\subsubsection{Overall performance}
On dataset $\mathcal{B}^{(w1)}$, \textsc{DiffPool} achieves the best performance among all baseline methods. It scores an Acc. of $72.12\%$ for predicting $\mathcal{A}^{(gen)}$, an Acc. of $25.87\%$ for predicting $\mathcal{A}^{(inc)}$, and an RMSE of $13.81$ for predicting $\mathcal{A}^{(age)}$.
While on dataset $\mathcal{B}^{(w2)}$, \textsc{SAGPool} and \textsc{DiffPool} give comparable best performances.
\textsc{SAGPool} slightly outperforms \textsc{DiffPool} that it scores a higher Acc.
for predicting $\mathcal{A}^{(inc)}$, and a lower RMSE 
for predicting $\mathcal{A}^{(age)}$.
Our proposed \textsc{CalendarGNN} outperforms all baseline methods across almost all metrics. On $\mathcal{B}^{(w1)}$, \textsc{CalendarGNN} scores an Acc. of $72.98\%$ for predicting $\mathcal{A}^{(gen)}$ ($+1.19\%$ relatively over \textsc{DiffPool}), an Acc. of $28.83\%$ for predicting $\mathcal{A}^{(inc)}$ ($+11.44\%$ relatively over \textsc{DiffPool}), and an RMSE of $13.60$ for $\mathcal{A}^{(age)}$ ($-1.52\%$ relatively over \textsc{DiffPool}).
On $\mathcal{B}^{(w2)}$, it scores an Acc. of $71.63\%$, an Acc. of $27.10\%$, and an RMSE of $13.88$ for predicting $\mathcal{A}^{(gen)}$, $\mathcal{A}^{(inc)}$, and $\mathcal{A}^{(age)}$ respectively ($+0.86\%$, $+10.52\%$, and $-2.32\%$  over \textsc{SAGPool}).
\textsc{CalendarGNN-Attn} further improves the Acc. for predicting $\mathcal{A}^{(inc)}$ to $29.67\%$ and $28.17\%$ on both datasets ($+2.9\%$ and $+3.9\%$ relatively over \textsc{CalendarGNN}); and, decreases RMSE for $\mathcal{A}^{(age)}$ to $13.52$ and $13.67$ ($-0.6\%$ and $-1.5\%$ relatively over \textsc{CalendarGNN}).

\subsubsection{Compare against behavior modeling methods}
\textsc{SR-GNN} gives the best performance of predicting user gender $\mathcal{A}^{(gen)}$ and user income $\mathcal{A}^{(inc)}$ among all behavior modeling methods. 
\textsc{LearnSuc} gives the best performance of predicting user age $\mathcal{A}^{(age)}$.
This is probably because \textsc{SR-GNN} learns embedding for sessions instead of users and inferring user age of real values based on session embeddings are difficult than directly using user embedding.
Beside, \textsc{SR-GNN} is designed to model session as a graph of items, but it ignores all spatial and temporal signals. On the contrary, our \textsc{CalendarGNN} models each user's behaviors as a single tripartite graph of sessions, locations, and items attributed by temporal signals. And, this user spatiotemporal behavior graph is able to capture the complex behavioral spatial and temporal patterns.
\textsc{CalendarGNN} outperforms \textsc{SR-GNN} by $+4.53\%$ and $+42.65\%$ relatively for Accs. of predicting $\mathcal{A}^{(gen)}$ and $\mathcal{A}^{(inc)}$ on dataset $\mathcal{B}^{(w1)}$, and by $+5.65\%$ and $+36.9\%$ on dataset $\mathcal{B}^{(w2)}$. \textsc{CalendarGNN} outperforms \textsc{LearnSuc} by $-12.20\%$ and $-18.02\%$ for the RMSEs of predicting $\mathcal{A}^{(age)}$.

\begin{figure}[t]
    \centering
    \subfigure[Clustering of user embeddings $\mathbf{u}$ is highly indicative about gender $\mathcal{A}^{(gen)}$]{
    	\includegraphics[width=0.475\linewidth]{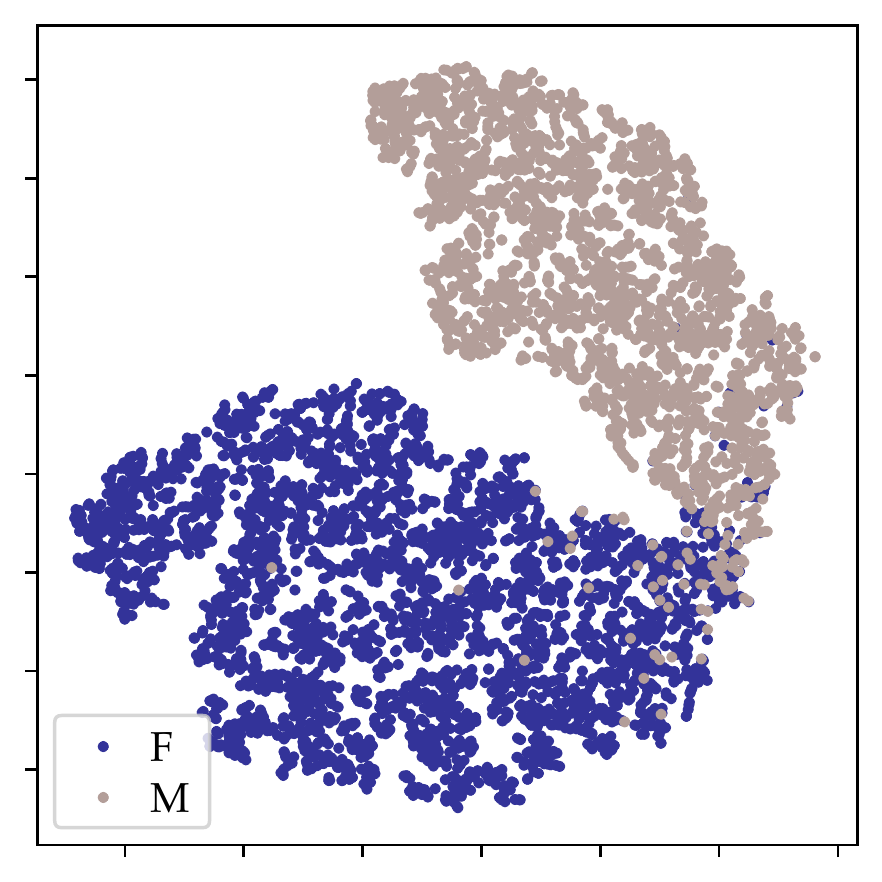}
		\label{fig:clustering1}}
    \hfill
    \subfigure[Clustering of spatial patterns $\mathbf{p}_{\mathcal{L}}$ is highly indicative about income $\mathcal{A}^{(inc)}$]{
    	\includegraphics[width=0.475\linewidth]{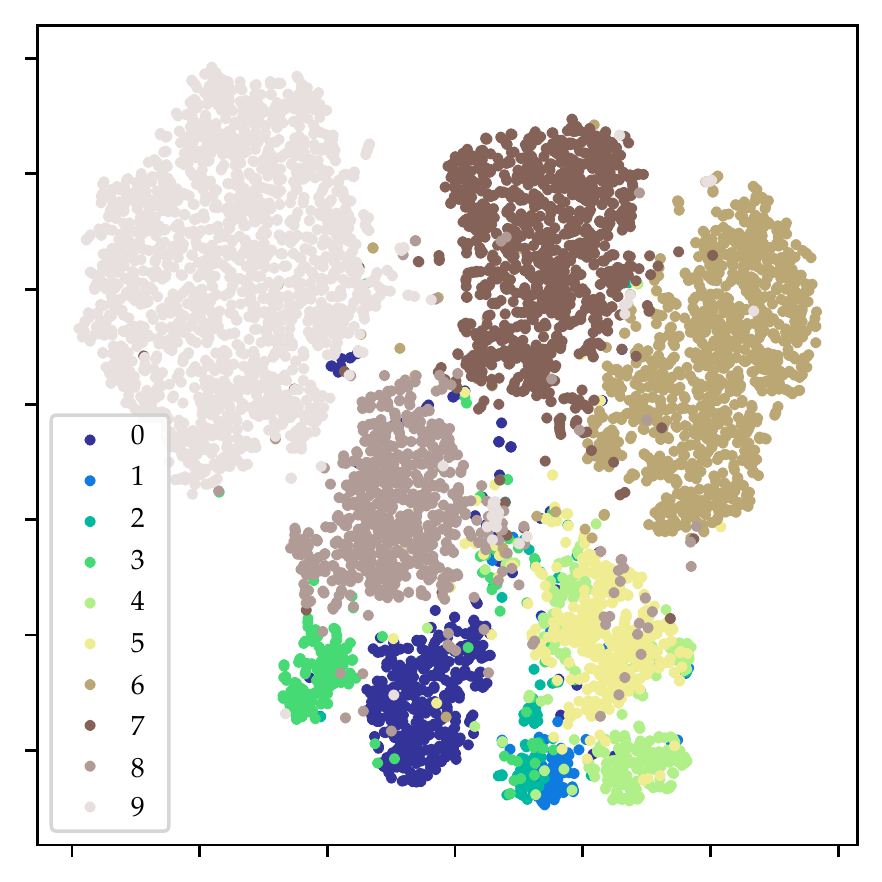}
		\label{fig:clustering2}}
	\vspace{-0.15in}
    \caption{Clustering of user embeddings and patterns}
    \label{fig:clustering}
    \vspace{-0.2in}
\end{figure}

\subsubsection{Compare against GNN methods}
\textsc{DiffPool} performs the best among all \textsc{GNN}-based baseline methods on dataset $\mathcal{B}^{(w1)}$. It scores an Acc. of $72.12\%$ for predicting user gender $\mathcal{A}^{(gen)}$ ($+3.29\%$ relatively over \textsc{SR-GNN}), an Acc. of $25.87\%$ for predicting user income $\mathcal{A}^{(inc)}$ ($+28.01\%$ relatively over \textsc{SR-GNN}), and an RMSE of $13.81$ for predicting user age $\mathcal{A}^{(age)}$ ($-10.85\%$ relatively over \textsc{LearnSuc}).
\textsc{SAGPool} shows competitive good performance on dataset $\mathcal{B}^{(w2)}$.
Both of these two methods learn hierarchical representation of general graphs. They are not designed to capture the specific tripartite graph structure of sessions, items, and locations. And, these methods are not capable of modeling the explicit time structures in user's spatiotemporal behaviors.

\textsc{DGCNN} underperforms \textsc{DiffPool} and \textsc{SAGPool} across all metrics on both datasets.
One reason is that \textsc{DGCNN}'s core component \textsc{SortPooling} layer relies on a node sorting algorithm (in analogous to sort continuous \textsc{WL} colors \cite{weisfeiler1968reduction}). This strategy produces lower performance for predicting user demographic labels compared with the learned hierarchical representations adopted by \textsc{DiffPool} and \textsc{SAGPool}.
\textsc{ECC} and \textsc{CapsGNN} yield slightly better performance than behavior modeling method \textsc{SR-GNN} for predicting user gender $\mathcal{A}^{(gen)}$. But, they can quite outperform \textsc{SR-GNN} for predicting $\mathcal{A}^{(inc)}$,
and outperform \textsc{LearnSuc} by a large margin for predicting $\mathcal{A}^{(age)}$.
This validates the spatiotemporal behavior graph of sessions, items, and locations (instead of itemset or simple item-session graph) provides more information for the GNN model.

Our \textsc{CalendarGNN} performs the best among all GNN-based methods across almost all metrics. On dataset $\mathcal{B}^{(w1)}$, \textsc{CalendarGNN} scores an Acc. of $72.98\%$ for $\mathcal{A}^{(gen)}$ ($+1.19\%$ relatively over \textsc{DiffPool}), an Acc. of $28.83\%$ for $\mathcal{A}^{(inc)}$ ($+11.44\%$ relatively over \textsc{DiffPool}), and an RMSE of $13.60$ for $\mathcal{A}^{(age)}$ ($-1.52\%$ relatively over \textsc{DiffPool}).
On dataset $\mathcal{B}^{(w2)}$, it scores an Acc. of $71.63\%$, an Acc. of $27.10\%$, and an RMSE of $13.88$ for predicting $\mathcal{A}^{(gen)}$, $\mathcal{A}^{(inc)}$, and $\mathcal{A}^{(age)}$ respectively ($+0.86\%$, $+10.52\%$, and $-2.32\%$  over \textsc{SAGPool}). This confirms that the proposed calendar-like neural architecture of \textsc{CalendarGNN} is able to distill user embeddings of greater predictive power on demographic labels.

By considering the interactions between spatial and temporal pattern, \textsc{CalendarGNN-Attn} further improves the Acc. for predicting $\mathcal{A}^{(inc)}$ to $29.67\%$ and $28.17\%$ on both datasets ($+2.9\%$ and $+3.9\%$ relatively over \textsc{CalendarGNN}); and, decreases RMSE for $\mathcal{A}^{(age)}$ to $13.52$ and $13.67$ ($-0.6\%$ and $-1.5\%$ relatively over \textsc{CalendarGNN}). We also note that \textsc{CalendarGNN-Attn} underperforms \textsc{CalendarGNN} on both datasets for predicting $\mathcal{A}^{(age)}$. This indicates the interactions between spatial and temporal patterns provide no extra information for predicting user genders. 
More results for examining the importance of each spatial or temporal pattern in different predictive tasks can be found in the supplemental materials

\subsection{Qualitative analysis}
In Figure \ref{fig:clustering}, we provide visualizations of user embeddings and patterns learned by \textsc{CalendarGNN} using \text{t-SNE} \cite{maaten2008visualizing}. The clustering results presented in Figure \ref{fig:clustering1} clearly demonstrate that the learned user embeddings are highly indicative about the target user attribute $\mathcal{A}^{(gen)}$. Furthermore, we plot the learned spatial patterns $\mathbf{p}_{\mathcal{L}}$ in Figure \ref{fig:clustering2} and it can be seen that they are especially useful for determining user's income level  $\mathcal{A}^{(gen)}$: users of high income levels (e.g., ``7'', ``8'' and ``9'') forms distinct non-overlapping clusters despite some users of lower income level (e.g., ``1'') and unknown (``0'') scatters at the bottom part. 

\begin{figure}[t]
    \centering
    \vspace{-0.1in}
    \subfigure[Change of $F1$ scores for predicting user gender $\mathcal{A}^{(gen)}$ on $\mathcal{B}^{(w1)}$]{
    	\includegraphics[width=0.425\linewidth]{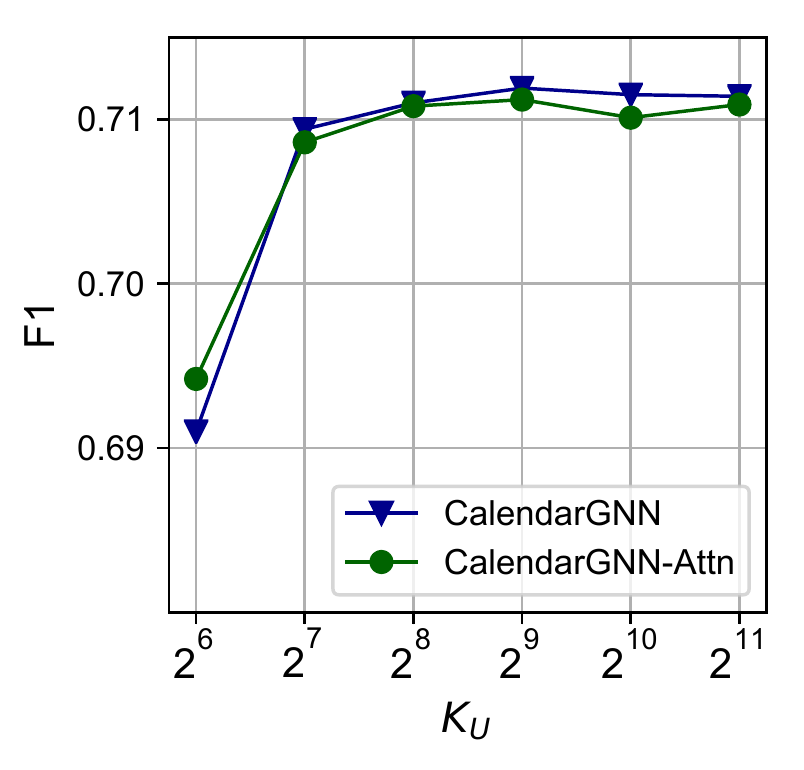}
		\label{fig:sensitivity}}
    \hfill
    \subfigure[Per epoch training time  w.r.t. average input graph size $\vert G \vert$]{
    	\includegraphics[width=0.425\linewidth]{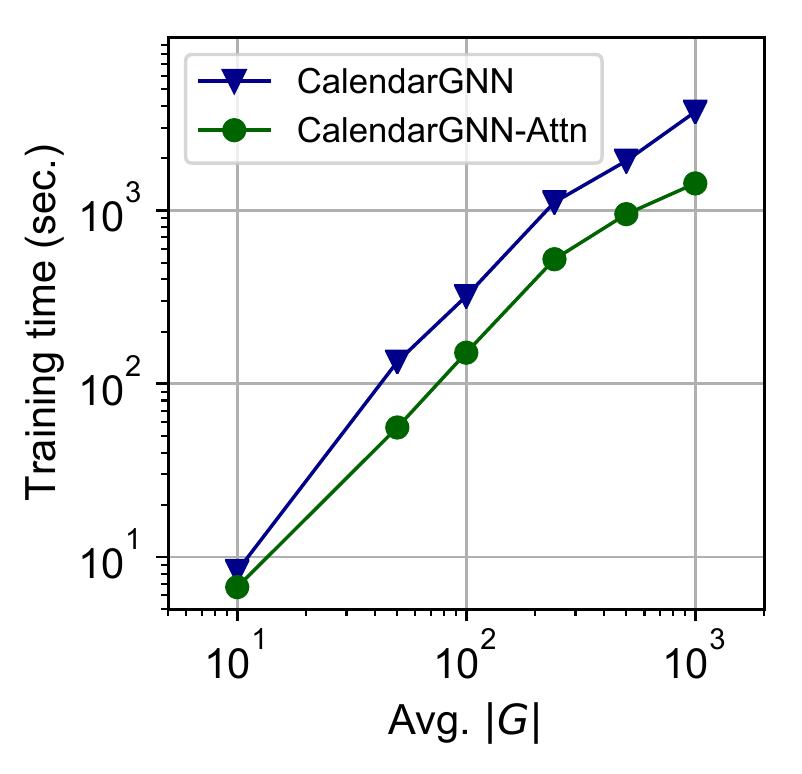}
		\label{fig:efficiency}}
	\vspace{-0.15in}
    \caption{Sensitivity and efficiency of \textsc{CalendarGNN}.}
    \vspace{-0.2in}
\end{figure}

\subsection{Sensitivity and Efficiency}
We test through \textsc{CalendarGNN}'s hyper-parameters. Figure \ref{fig:sensitivity} shows the prediction performance is stable for a range of user embedding dimensions $K_{\mathcal{U}}$ from $2^{7}$ to $2^{11}$. We also test the model's efficiency. All experiments are conducted on single server with dual 12-core Intel Xeon 2.10GHz CPUs with single NVIDIA GeForce GTX 2080 Ti GPU. Figure \ref{fig:efficiency} shows the per epoch training time is linear to the average size of input user spatiotemporal graphs.

%% file: 6conclusions.tex

In this work, we proposed a novel Graph Neural Network (GNN) model for learning user representations from spatiotemporal behavior data. It aggregates embeddings of items and locations into session embeddings, and generates user embedding on the calendar neural architecture.
Experiments on two real datasets demonstrate the effectiveness of our method.